\documentclass{article}

\usepackage{microtype}
\usepackage{graphicx}
\usepackage{booktabs} % for professional tables
\usepackage{amsfonts}
\usepackage{amsmath}
\usepackage{subfig}
\usepackage[dvipsnames]{xcolor}

\newtheorem{note}{Note}
\newtheorem{theorem}{Theorem}

\usepackage{hyperref}

\usepackage[accepted]{icml2020}

\icmltitlerunning{Unraveling Meta-Learning}

\begin{document}

\twocolumn[
\icmltitle{Unraveling Meta-Learning: Understanding Feature\\Representations for Few-Shot Tasks}

% It is OKAY to include author information, even for blind
% submissions: the style file will automatically remove it for you
% unless you've provided the [accepted] option to the icml2020
% package.

% List of affiliations: The first argument should be a (short)
% identifier you will use later to specify author affiliations
% Academic affiliations should list Department, University, City, Region, Country
% Industry affiliations should list Company, City, Region, Country

% You can specify symbols, otherwise they are numbered in order.
% Ideally, you should not use this facility. Affiliations will be numbered
% in order of appearance and this is the preferred way.
\icmlsetsymbol{equal}{*}

\begin{icmlauthorlist}
\icmlauthor{Micah Goldblum}{umd}
\icmlauthor{Steven Reich}{equal,umd}
\icmlauthor{Liam Fowl}{equal,umd}
\icmlauthor{Renkun Ni}{equal,umd}
\icmlauthor{Valeriia Cherepanova}{equal,umd}
\icmlauthor{Tom Goldstein}{umd}
\end{icmlauthorlist}

\icmlaffiliation{umd}{University of Maryland, College Park}

\icmlcorrespondingauthor{Micah Goldblum}{goldblum@umd.edu}

% You may provide any keywords that you
% find helpful for describing your paper; these are used to populate
% the "keywords" metadata in the PDF but will not be shown in the document
\icmlkeywords{Few-Shot, Meta-Learning, Machine Learning, ICML}

\vskip 0.3in
]

% this must go after the closing bracket ] following \twocolumn[ ...

% This command actually creates the footnote in the first column
% listing the affiliations and the copyright notice.
% The command takes one argument, which is text to display at the start of the footnote.
% The \icmlEqualContribution command is standard text for equal contribution.
% Remove it (just {}) if you do not need this facility.

%\printAffiliationsAndNotice{}  % leave blank if no need to mention equal contribution
\printAffiliationsAndNotice{\icmlEqualContribution} % otherwise use the standard text.

\begin{abstract}
Meta-learning algorithms produce feature extractors which achieve state-of-the-art performance on few-shot classification.  While the literature is rich with meta-learning methods, little is known about why the resulting feature extractors perform so well.  We develop a better understanding of the underlying mechanics of meta-learning and the difference between models trained using meta-learning and models which are trained classically.  In doing so, we introduce and verify several hypotheses for why meta-learned models perform better.  Furthermore, we develop a regularizer which boosts the performance of standard training routines for few-shot classification.  In many cases, our routine outperforms meta-learning while simultaneously running an order of magnitude faster.
\end{abstract}

\section{Introduction}
\label{Introduction}

\par Training neural networks from scratch requires large amounts of labeled data, making it impractical in many settings. 
%In applications like drug discovers \cite{altae2017low} where data is expensive and time consuming to obtain. 
When data is expensive or time consuming to obtain, training from scratch may be cost prohibitive \cite{altae2017low}.
In other scenarios, models must adapt efficiently to changing environments before enough time has passed to amass a large and diverse data corpus \cite{nagabandi2018learning}.  In both of these cases, massive state-of-the-art networks would overfit to the tiny training sets available.  To overcome this problem, practitioners pre-train on large auxiliary datasets and then fine-tune the resulting models on the target task.  For example, ImageNet pre-training of large ResNets has become an industry standard for transfer learning \cite{kornblith2019better}.  Unfortunately, transfer learning from classically trained models often yields sub-par performance in the extremely data-scarce regime or breaks down entirely when only a few data samples are available in the target domain. 

\par Recently, a number of few-shot benchmarks have been rapidly improved using {\em meta-learning} methods  \cite{lee2019meta, song2019fast}.  Unlike classical transfer learning, which uses a base model pre-trained on a different task, meta-learning algorithms produce a base network that is specifically designed for quick adaptation to new tasks using few-shot data. Furthermore, meta-learning is still effective when applied to small, lightweight base models that can be fine-tuned with relatively few computations.

%The framework for meta-learning entails finding parameters from which the network adapts effectively to new tasks.  We provide a thorough treatment of the meta-learning framework in the following section.

\par The ability of meta-learned networks to rapidly adapt to new domains suggests that {\em the feature representations learned by meta-learning must be fundamentally different than feature representations learned through conventional training.}  Because of the good performance that meta-learning offers in various settings, many researchers have been content to use these features without considering how or why they differ from conventional representations.  As a result, little is known about the fundamental differences between meta-learned feature extractors and those which result from classical training.  Training routines are often treated like a black box in which high performance is celebrated, but a deeper understanding of the phenomenon remains elusive.  To further complicate matters, a myriad of meta-learning strategies exist that may exploit different mechanisms.

\par In this paper, we delve into the differences between features learned by meta-learning and classical training.  We explore and visualize the behaviors of different methods and identify two different mechanisms by which meta-learned representations can improve few-shot learning.  In the case of meta-learning strategies that fix the feature extractor and only update the last (classification) layer of a network during the inner-loop, such as MetaOptNet \cite{lee2019meta} and R2-D2 \cite{bertinetto2018meta}, we find that meta-learning tends to cluster object classes more tightly in feature space.  As a result, the classification boundaries learned during fine-tuning are less sensitive to the choice of few-shot samples.  In the second case, we hypothesize that meta-learning strategies that use end-to-end fine-tuning, such as Reptile \cite{nichol2018reptile}, search for meta-parameters that lie close in weight space to a wide range of task-specific minima.  In this case, a small number of SGD steps can transport the parameters to a good minimum for a specific task.

\par Inspired by these observations, we propose simple regularizers that improve feature space clustering and parameter-space proximity.  These regularizers boost few-shot performance appreciably, and improving feature clustering does so without the dramatic increase in optimization cost that comes from conventional meta-learning.

% We study traits of different meta-learning algorithms, and we develop two hypotheses that explain performance benefits over transfer learning for few-shot tasks: clustering in feature space and clustering in parameter space.

% \begin{itemize}
%     \item We first study the feature space of meta-learning.  We find that linear separability of features is not enough for few-shot learning, and low variation of features within classes compared to variation between classes is important for few-shot performance.  This observation leads us to develop two regularizers for transfer learning which consistently improve the performance of classically trained models on few-shot tasks.
%     \item Second, we observe that Reptile resembles a consensus optimization algorithm.  We thus hypothesize that this method finds parameters which lie in close proximity to minima in the loss landscape of few-shot tasks.  We then introduce a regularizer to encourage this phenomenon in Reptile, and we improve performance on both 1-shot and 5-shot mini-ImageNet.
% \end{itemize}

\section{Problem Setting}
\label{ProblemSetting}

\subsection{The Meta-Learning Framework}
\label{MetaLearning}

\par In the context of few-shot learning, the objective of meta-learning algorithms is to produce a network that quickly adapts to new classes using little data. Concretely stated, meta-learning algorithms find parameters that can be fine-tuned in few optimization steps and on few data points in order to achieve good generalization on a task $\mathcal{T}_i$, consisting of a small number of data samples from a distribution and label space that was not seen during training. The task is characterized as \textit{n-way}, \textit{k-shot} if the meta-learning algorithm must adapt to classify data from $\mathcal{T}_i$ after seeing $k$ examples from each of the $n$ classes in $\mathcal{T}_i$.

\par Meta-learning schemes typically rely on bi-level optimization problems with an \textit{inner loop} and an \textit{outer loop}.  An iteration of the outer loop involves first sampling a ``task,'' which comprises two sets of labeled data: the support data, $\mathcal{T}_i^s$, and the query data, $\mathcal{T}_i^q$. Then, in the inner loop, the model being trained is fine-tuned using the support data. Finally, the routine moves back to the outer loop, where the meta-learning algorithm minimizes loss on the query data with respect to the pre-fine-tuned weights.  This minimization is executed by differentiating through the inner loop computation and updating the network parameters to make the inner loop fine-tuning as effective as possible. Note that, in contrast to standard transfer learning (which uses classical training and simple first-order gradient information to update parameters), meta-learning algorithms differentiate through the entire fine-tuning loop.  A formal description of this process can be found in Algorithm \ref{alg:MetaAlgorithm}, as seen in \cite{goldblum2019robust}.

\begin{algorithm}[h]
   \caption{The meta-learning framework}
   \label{alg:MetaAlgorithm}
\begin{algorithmic}
\STATE {\bfseries Require:} Base model, 
 $F_\theta$, fine-tuning algorithm, $A$, learning rate, $\gamma$, and distribution over tasks, $p(\mathcal{T})$.
\STATE Initialize $\theta$, the weights of $F$; \\
\WHILE{not done}
\STATE Sample batch of tasks, $\{\mathcal{T}_i\}_{i=1}^n$, where $\mathcal{T}_i \sim p(\mathcal{T})$ and $\mathcal{T}_i = (\mathcal{T}_i^s, \mathcal{T}_i^q)$. \\
\FOR{$i=1,\dots,n$}
\STATE Fine-tune model on $\mathcal{T}_i$ (inner loop). New network parameters are written $\theta_{i} = A(\theta, \mathcal{T}_i^s)$. \\
\STATE Compute gradient $g_i = \nabla_{\theta} \mathcal{L}(F_{\theta_{i}}, \mathcal{T}_i^{q})$
\ENDFOR
\STATE Update base model parameters (outer loop): \\ $\theta \leftarrow \theta - \frac{\gamma}{n} \sum_i g_i$
\ENDWHILE
\end{algorithmic}
\end{algorithm}

\subsection{Meta-Learning Algorithms}
\par A variety of meta-learning algorithms exist, mostly differing in how they fine-tune on support data during the inner loop.  Some meta-learning approaches, such as MAML, update all network parameters using gradient descent during fine-tuning \cite{finn2017model}.  Because differentiating through the inner loop is memory and computationally intensive, the fine-tuning process consists of only a few (sometimes just 1) SGD steps.

\par Reptile, which functions as a zero'th-order approximation to MAML, avoids unrolling the inner loop and differentiating through the SGD steps.  Instead, after fine-tuning on support data, Reptile moves the central parameter vector in the direction of the fine-tuned parameters during the outer loop \cite{nichol2018reptile}.  In many cases, Reptile achieves better performance than MAML without having to differentiate through the fine-tuning process.

\par Another class of algorithms freezes the feature extraction layers during the inner loop; only the linear classifier layer is trained during fine-tuning.  Such methods include R2-D2 and MetaOptNet  \cite{bertinetto2018meta, lee2019meta}. The advantage of this approach is that the fine-tuning problem is now a convex optimization problem. Unlike MAML, which simulates the fine-tuning process using only a few gradient updates, last-layer meta-learning methods can use differentiable optimizers to exactly minimize the fine-tuning objective and then differentiate the solution with respect to feature inputs. Moreover, differentiating through these solvers is computationally cheap compared to MAML's differentiation through SGD steps on the whole network. While MetaOptNet relies on an SVM loss, R2-D2 simplifies the process even further by using a quadratic objective with a closed-form solution.  R2-D2 and MetaOptNet achieve stronger performance than MAML and are able to harness larger architectures without overfitting. 

\begin{table*}[h]
\begin{center}
\begin{tabular}{|c|c|c|c|c|}
\hline
Model & SVM & RR & ProtoNet & MAML \\ \hline
MetaOptNet-M & \textbf{62.64} $\pm$ 0.31 \%& \textbf{60.50} $\pm$ 0.30 \%& \textbf{51.99} $\pm$ 0.33 \%& \textbf{55.77} $\pm$ 0.32 \%\\ %\hline
MetaOptNet-C & 56.18 $\pm$ 0.31 \%& 55.09 $\pm$ 0.30 \%& 41.89 $\pm$ 0.32 \%& 46.39 $\pm$ 0.28 \%\\ \hline
R2-D2-M & \textbf{51.80} $\pm$ 0.20 \%& \textbf{55.89} $\pm$ 0.31 \%& \textbf{47.89} $\pm$ 0.32 \%& \textbf{53.72} $\pm$ 0.33 \%\\ %\hline
R2-D2-C & 48.39 $\pm$ 0.29 \%& 48.29 $\pm$ 0.29 \%& 28.77 $\pm$ 0.24 \%& 44.31 $\pm$ 0.28 \%\\ \hline
\end{tabular}
\end{center}
\caption{Comparison of meta-learning and classical transfer learning models with various fine-tuning algorithms on 1-shot mini-ImageNet.  ``MetaOptNet-M'' and ``MetaOptNet-C'' denote models with MetaOptNet backbone trained with MetaOptNet-SVM and classical training.  Similarly, ``R2-D2-M'' and ``R2-D2-C'' denote models with R2-D2 backbone trained with ridge regression (RR) and classical training.  Column headers denote the fine-tuning algorithm used for evaluation, and the radius of confidence intervals is one standard error.}
\label{MetaVsTransfer}
\end{table*}

\par Another last-layer method, ProtoNet, classifies examples by the proximity of their features to those of class centroids - a metric learning approach - in its inner loop \cite{snell2017prototypical}.  Again, the feature extractor's parameters are frozen in the inner loop, and the extracted features are used to create class centroids which then determine the network's class boundaries.  Because calculating class centroids is mathematically simple, this algorithm is able to efficiently backpropagate through this calculation to adjust the feature extractor. 

\par In this work, ``classically trained'' models are trained, using cross-entropy loss and SGD, on all classes simultaneously, and the feature extractors are adapted to new tasks using the same fine-tuning procedures as the meta-learned models for fair comparison.  This approach represents the industry-standard method of transfer learning using pre-trained feature extractors.

\subsection{Few-Shot Datasets}
\label{Datasets}

\par Several datasets have been developed for few-shot learning. We focus our attention on two datasets: mini-ImageNet and CIFAR-FS. Mini-ImageNet is a pruned and downsized version of the ImageNet classification dataset, consisting of 60,000, $84 \times 84$ RGB color images from $100$ classes \cite{vinyals2016matching}. These 100 classes are split into $64, 16,$ and $20$ classes for training, validation, and testing sets, respectively. The CIFAR-FS dataset samples images from CIFAR-100 \cite{bertinetto2018meta}. CIFAR-FS is split in the same way as mini-ImageNet with 60,000 $32 \times 32$ RGB color images from $100$ classes divided into $64, 16,$ and $20$ classes for training, validation, and testing sets, respectively.  

\subsection{Related Work}
\label{RelatedWork}

\iffalse
\begin{itemize}
    \item ``A Closer Look at Few-shot Classification''
    \item ``A Baseline for Few-Shot Image Classification''
    \item ``Pay Attention to Features, Transfer Learn faster CNNs''
    \item Any other work which attempts to understand meta-learning
\end{itemize}

\fi

\par In addition to introducing new methods for few-shot learning, recent work has increased our understanding of why some models perform better than others at few-shot tasks.  One such exploration performs baseline testing and discovers that network size has a large effect on the success of meta-learning algorithms \cite{chen2019closer}. Specifically, on some very large architectures, the performance of transfer learning approaches that of some meta-learning algorithms.  We thus focus on architectures common in the meta-learning literature.  Methods for improving transfer learning in the few-shot classification setting focus on much larger backbone networks \cite{chen2019closer, dhillon2019baseline}.  

\par Other work on transfer learning has found that feature extractors trained on large complex tasks can be more effectively deployed in a transfer learning setting by distilling knowledge about only important features for the transfer task \cite{wang2020pay}. Yet other work finds that features generated by a pre-trained model on data from classes absent from training are entangled, but the logits of the unseen data tend to be clustered \cite{frosst2019analyzing}.  Meta-learners without supervision in the outer loop have been found to perform well when equipped with a clustering-based penalty in the meta-objective \cite{huang2019centroid}. Work on standard supervised learning has alternatively studied low-dimensional structures via rank \cite{goldblum2019truth, sainath2013low}.

\par While improvements have been made to meta-learning algorithms and transfer learning approaches to few-shot learning, little work has been done on understanding the underlying mechanisms that cause meta-learning routines to perform better than classically trained models in data scarce settings.

\section{Are Meta-Learned Features Fundamentally Better for Few-Shot Learning?}
\label{PerformBetter}

\par It has been said that meta-learned models ``learn to learn'' \cite{finn2017model}, but one might ask if they instead learn to optimize; their features could simply be well-adapted for the specific fine-tuning optimizers on which they are trained.  We dispel the latter notion in this section.

\par In Table \ref{MetaVsTransfer}, we test the performance of meta-learned feature extractors not only with their own fine-tuning algorithm, but with a variety of fine-tuning algorithms. We find that in all cases, the meta-learned feature extractors outperform classically trained models of the same architecture. See Appendix \ref{sec:ModelSwap} for results from additional experiments. 

\par This performance advantage across the board suggests that meta-learned features are qualitatively different than conventional features and fundamentally superior for few-shot learning. The remainder of this work will explore the characteristics of meta-learned models.

\iffalse

\par To further highlight the fundamental differences between meta-learned and classically trained networks, we contrast their feature representations.  Since we hypothesize that meta-learned models have clustering tendencies in feature space, one might expect that meta-learned models exhibit sparse saliency maps.  Figure [REFERENCE FIGURE] shows that indeed, meta-learned models use sparser features in the data.  These saliency maps are generated by [INSERT MODELS] on [INSERT DATA].

\par [INSERT SALIENCY MAPS]

\fi

%\par We conclude from these brief experiments that meta-learned models are fundamentally better at few-shot tasks than their classically trained counterparts.  

\section{Class Clustering in Feature Space}
\label{FeatureSpace}

\par Methods such as ProtoNet, MetaOptNet, and R2-D2 fix their feature extractor during fine-tuning.  For this reason, they must learn to embed features in a way that enables few-shot classification.  For example, MetaOptNet and R2-D2 require that classes are linearly separable in feature space, but mere linear separability is not a sufficient condition for good few-shot performance.  The feature representations of randomly sampled few-shot data from a given class must not vary so much as to cause classification performance to be sample-dependent.  In this section, we examine clustering in feature space, and we find that meta-learned models separate features differently than classically trained networks.

\par  %In Sections \ref{Regularizer1} and \ref{HyperplaneInvariance}, we will introduce two regularizers, $R_{FC}$ and $R_{HV}$, which encourage class separation and independence of decision boundaries from data sampling, respectively.  

\subsection{Measuring Clustering in Feature Space}

\par We begin by measuring how well different training methods cluster feature representations. To measure feature clustering (FC), we consider the intra-class to inter-class variance ratio 
$$\frac{\sigma_{within}^{2}}{\sigma_{between}^{2}}=\frac{C}{N}\frac{\sum_{i,j}\|\phi_{i,j}-\mu_i\|_2^2}{\sum_{i}\|\mu_{i}-\mu\|_2^2},$$
where $\phi_{i,j}$ is a feature vector in class $i$, $\mu_i$ is the mean of feature vectors in class $i$, $\mu$ is the mean across all feature vectors, $C$ is the number of classes, and $N$ is the number of data points per class.  Low values of this fraction correspond to collections of features such that classes are well-separated and a hyperplane formed by choosing a point from each of two classes does not vary dramatically with the choice of samples.

\par In Table \ref{RegValues}, we highlight the superior class separation of meta-learning methods.  We compute two quantities, $R_{FC}$ and $R_{HV}$, for MetaOptNet and R2-D2 as well as classical transfer learning baselines of the same architectures.  These two quantities measure the intra-class to inter-class variance ratio and invariance of separating hyperplanes to data sampling.  Mathematical formulations of $R_{FC}$ and $R_{HV}$ can be found in Sections \ref{Regularizer1} and \ref{HyperplaneInvariance}, respectively.  Lower values of each measurement correspond to better class separation.  On both CIFAR-FS and mini-ImageNet, the meta-learned models attain lower values, indicating that feature space clustering plays a role in the effectiveness of meta-learning.

\begin{table}[h!]
\begin{center}
\begin{tabular}{|c|c|c|c|c|}
\hline
Training & Dataset & $R_{FC}$ & $R_{HV}$ \\ \hline
R2-D2-M & CIFAR-FS& {\bf 1.29} & {\bf 0.95} \\ %\hline
R2-D2-C & CIFAR-FS& 2.92 & 1.69 \\ \hline
MetaOptNet-M & CIFAR-FS & {\bf 0.99} & {\bf 0.75} \\ %\hline
MetaOptNet-C & CIFAR-FS& 1.84 & 1.25 \\ \hline
R2-D2-M & mini-ImageNet& {\bf 2.60} & {\bf 1.57} \\ %\hline
R2-D2-C & mini-ImageNet& 3.58 & 1.90 \\ \hline
MetaOptNet-M & mini-ImageNet & {\bf 1.29} & {\bf 0.95} \\ %\hline
MetaOptNet-C & mini-ImageNet& 3.13 & 1.75 \\ \hline

\end{tabular}
\end{center}
\caption{Comparison of class separation metrics for feature extractors trained by classical and meta-learning routines. $R_{FC}$ and $R_{HV}$ are measurements of feature clustering and hyperplane variation, respectively, and we formalize these measurements below.  In both cases, lower values correspond to better class separation.  We pair together models according to dataset and backbone architecture. ``-C'' and ``-M'' respectively denote classical training and meta-learning. See Sections \ref{Regularizer1} and \ref{HyperplaneInvariance} for more details.}
\label{RegValues}
\end{table}

\subsection{Why is Clustering Important?}
\par
\begin{figure}[h]
    \centering
    \subfloat[]{{\includegraphics[width=\columnwidth]{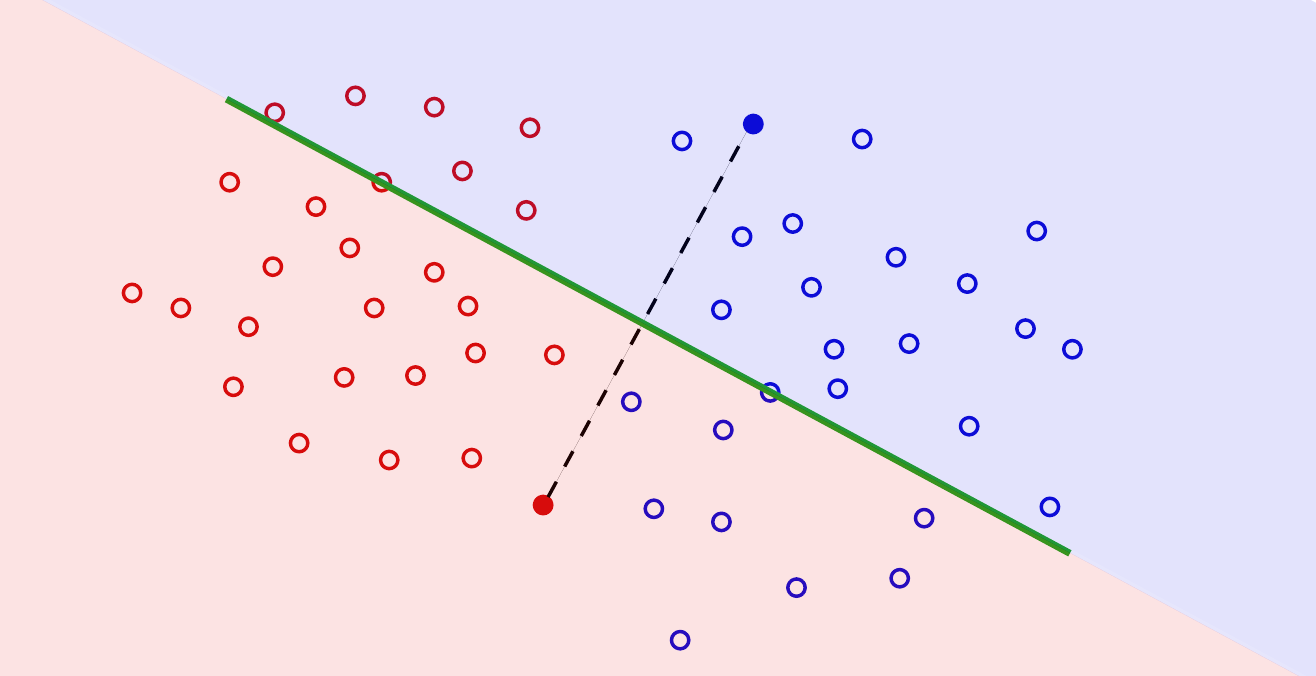} }}
    \qquad
    \subfloat[]{{\includegraphics[width=\columnwidth]{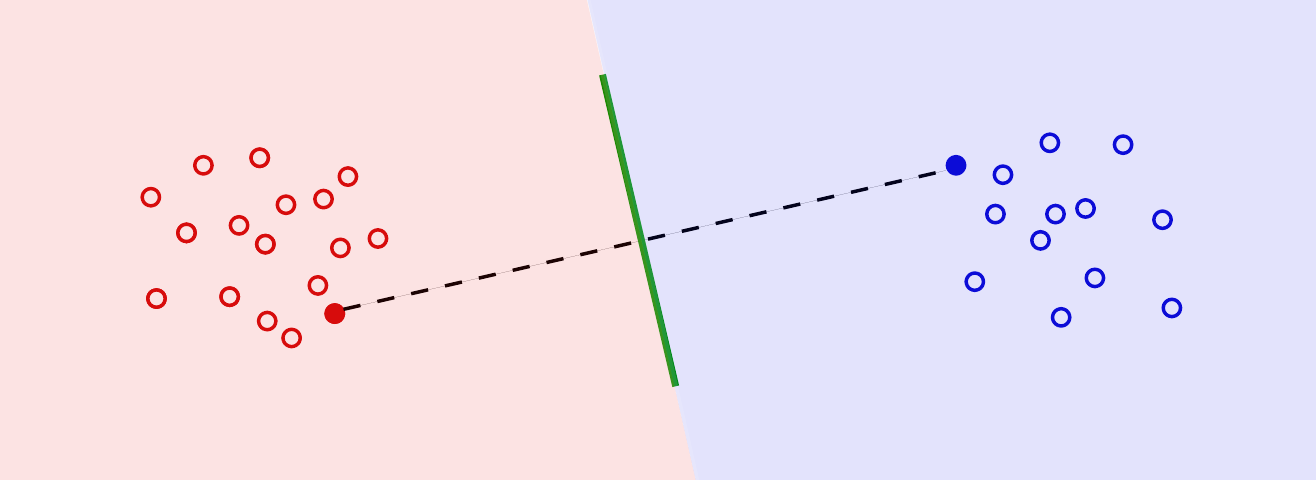} }}
    \caption{a) When class variation is high relative to the variation between classes, decision boundaries formed by one-shot learning are inaccurate, even though classes are linearly separable.  b) As classes move farther apart relative to the class variation, one-shot learning yields better decision boundaries.}
    \label{fig:toy_problem}
\end{figure}

 To demonstrate why linear separability is insufficient for few-shot learning, consider Figure \ref{fig:toy_problem}.  As features in a class become spread out and the classes are brought closer together, the classification boundaries formed by sampling one-shot data often misclassify large regions.  In contrast, as features in a class are compacted and classes move far apart from each other, the intra-class to inter-class variance ratio drops, and dependence of the class boundary on the choice of one-shot samples becomes weaker. 

This intuitive argument is formalized in the following result.
\begin{theorem}
Consider two random variables, $X$ representing class $1,$ and $Y$ representing class $2.$ Let $U$ be the random variable equal to $X$ with probability $1/2,$ and $Y$ with probability $1/2.$ Assume the variance ratio bound
$$ \frac{\text{Var}[X]+\text{Var}[Y]}{\text{Var}[U]}  <\epsilon$$
holds for sufficiently small $\epsilon \ge 0.$

Draw random one-shot data, $x\sim X$ and $y \sim Y,$  and a test point $z\sim X.$ Consider the linear classifier 
$$ c(z) = \begin{cases}
  1, \quad \text{if }  z^T(x - y) - \frac{1}{2}\|x\|^2 +\frac{1}{2}\|y\|^2 \ge 0\\
  2,\quad  \text{otherwise}.
 \end{cases}$$
This classifier assigns the correct label to $z$ with probability at least
$$1 - \frac{32\epsilon}{1-\epsilon}.$$
%$$1 - \frac{\epsilon}{1-\epsilon}\frac{32\text{Var}[X]+16\text{Var}[Y]}{(\text{Var}[X]+\text{Var}[Y])} .$$
\end{theorem}
Note that the linear classifier in the theorem is simply the maximum-margin linear classifier that separates the two training points. 
In plain words, Theorem 1 guarantees that one-shot learning performance is effective when the variance ratio is small, with classification becoming asymptotically perfect as the ratio approaches zero.  A proof is provided in Appendix \ref{sec:Proof}.

%\begin{figure}
%    \centering
%    \subfloat[]{{\includegraphics[width=5cm]{Images/LDA_close.pdf} }}
%    \qquad
%    \subfloat[]{{\includegraphics[width=5cm]{Images/LDA_far.pdf} }}
%    \caption{When class variation is high relative to the variation between classes, decision boundaries formed by one-shot learning are inaccurate (a).  As classes move farther apart relative to the class variation, one-shot learning yields better decision boundaries (b).}
%    \label{fig:toy_problem}
%\end{figure}

% \par Note that the ease of learning from clustered feature representations is leveraged by linear discriminant analysis (LDA), which projects data onto directions which minimize the intra-class to inter-class variance ratio \cite{mika1999fisher}.

\subsection{Comparing Feature Representations of Meta-Learning and Classically Trained Models}
\label{VisualizingFeatures}

\par
\begin{figure}[h]
    \centering
    %\subfloat[MAML]{{\includegraphics[width=\columnwidth]{Images/LDA_MAML_large.pdf} }}
    %\qquad
    \subfloat[Meta-Learning]{{\includegraphics[width=\columnwidth]{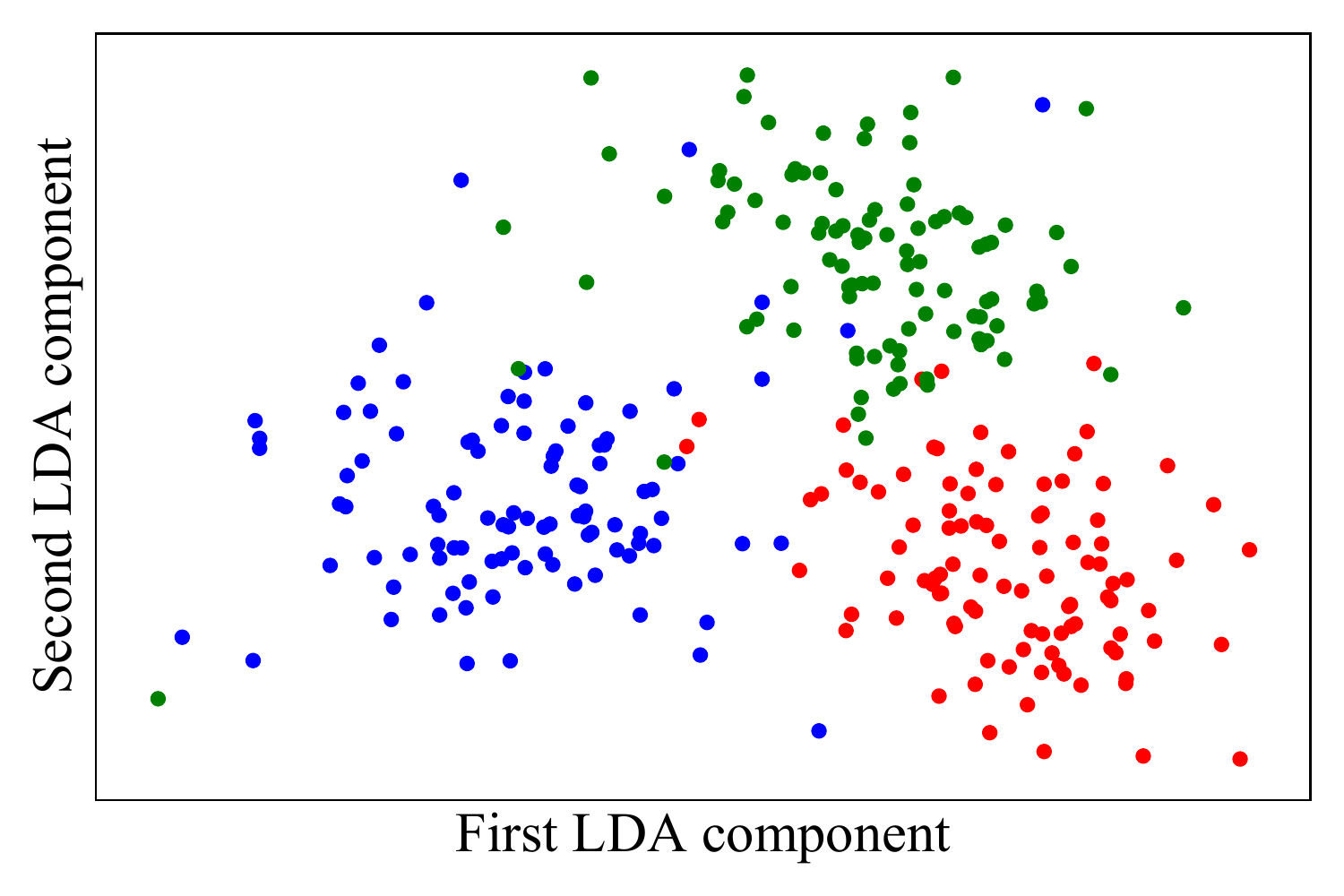} }}
    \qquad
    \subfloat[Classically Trained]{{\includegraphics[width=\columnwidth]{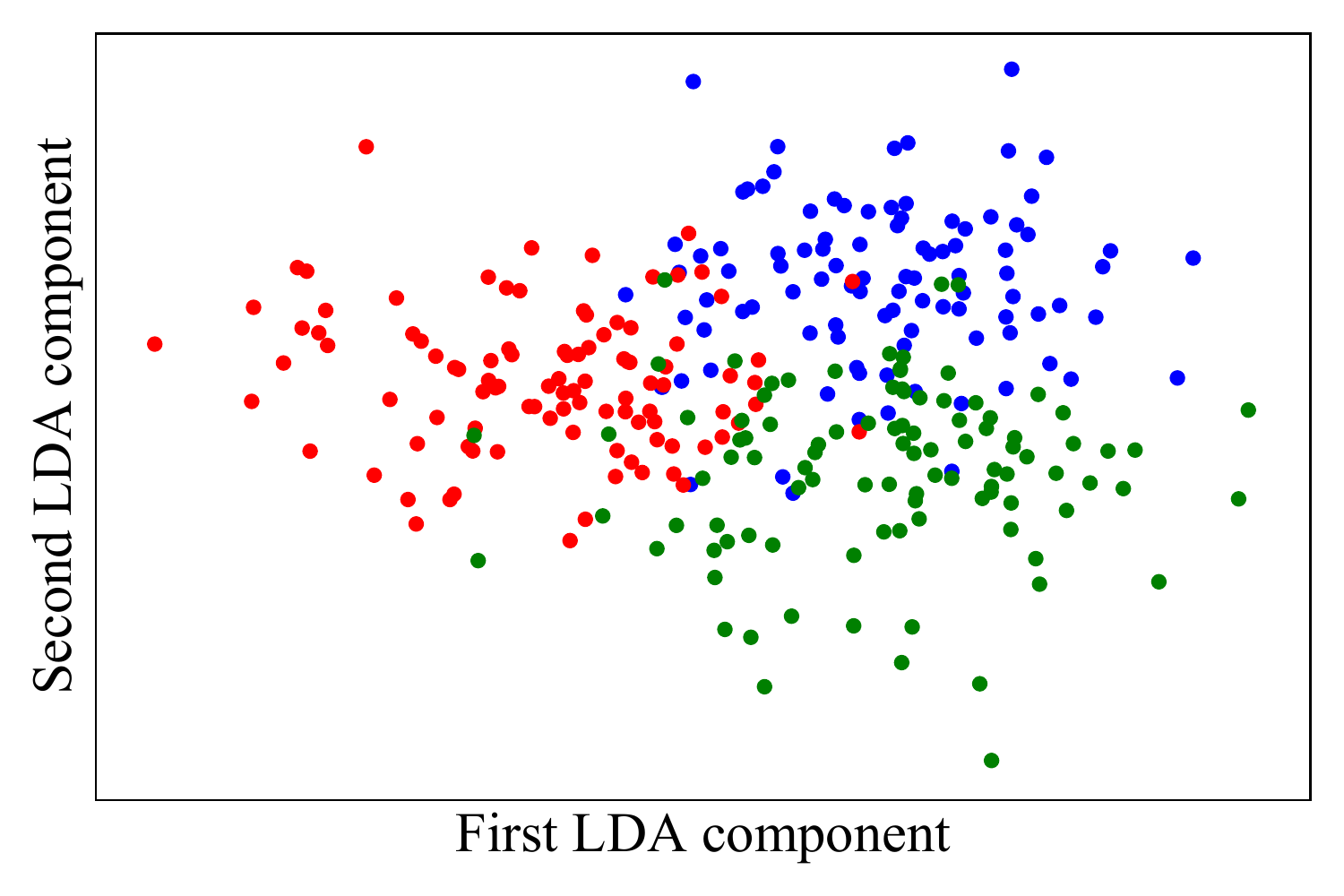} }}
    \caption{Features extracted from mini-ImageNet test data by a) ProtoNet and b) classically trained models with identical architectures (4 convolutional layers).  The meta-learned network produces better class separation.}
    \label{fig:LDA}
\end{figure}

 We begin our investigation into the feature space of meta-learned models by visualizing features.  Figure \ref{fig:LDA} contains a visual comparison of ProtoNet and a classically trained model of the same architecture on mini-ImageNet.  Three classes are randomly chosen from the test set, and $100$ samples are taken from each class.  The samples are then passed through the feature extractor, and the resulting vectors are plotted.  Because feature space is high-dimensional, we perform a linear projection into $\mathbb{R}^2$.  We project onto the first two component vectors determined by LDA.  Linear discriminant analysis (LDA) projects data onto directions that minimize the intra-class to inter-class variance ratio \cite{mika1999fisher}, and LDA is therefore ideal for visualizing the class separation phenomenon.

\par In the plots, we see that relative to the size of the point clusters, the classically trained model mashes features together, while the meta-learned models draws the classes farther apart.  While visually separate class features may be neither a necessary nor sufficient condition for few-shot performance, we take these plots as inspiration for our regularizer in the following section.  %Surprisingly, MAML, which updates all network parameters during fine-tuning, exhibits almost as good class separation as ProtoNet.

\subsection{Feature Space Clustering Improves the Few-Shot Performance of Transfer Learning}
\label{Regularizer1}

\par We now further test the feature clustering hypothesis by promoting the same behavior in classically trained models.  Consider a network with feature extractor $f_\theta$ and fully-connected layer $g_w$.  Then, denoting training data in class $i$ by $\{x_{i,j}\}$, we formulate the feature clustering regularizer by
$$R_{FC}(\theta, \{x_{i,j}\}) = \frac{C}{N} \frac{\sum_{i,j}\|f_\theta(x_{i,j})-\mu_i\|_2^2}{\sum_{i}\|\mu_{i}-\mu\|_2^2},
$$
where $f_\theta (x_{i,j})$ is a feature vector corresponding to a data point in class $i$, $\mu_i$ is the mean of feature vectors in class $i$, and $\mu$ is the mean across all feature vectors.  When this regularizer has value zero, classes are represented by distinct point masses in feature space, and thus the class boundary is invariant to the choice of few-shot data.

\par We incorporate this regularizer into a standard training routine by sampling two images per class in each mini-batch so that we can compute a within-class variance estimate.  Then, the total loss function becomes the sum of cross-entropy and $R_{FC}$.  We train the R2-D2 and MetaOptNet backbones in this fashion on the mini-ImageNet and CIFAR-FS datasets, and we test these networks on both 1-shot and 5-shot tasks.  In all experiments, feature clustering improves the performance of transfer learning and sometimes even achieves higher performance than meta-learning. Furthermore, the regularizer does not appreciably slow down classical training, which, without the expense of differentiating through an inner loop, runs as much as 13 times faster than the corresponding meta-learning routine.  See Table \ref{Feature_Space_Reg} for numerical results, and see Appendix \ref{sec:FeatureExpts} for experimental details including training times.

\begin{table*}[h!]
\begin{center}
\begin{tabular}{lcccccc}
\hline
 & &\multicolumn{2}{c}{mini-ImageNet}&\multicolumn{2}{c}{CIFAR-FS}\\\cline{3-6}
Training & Backbone & 1-shot & 5-shot & 1-shot & 5-shot\\ \hline
R2-D2 & R2-D2 & ${\bf 51.80}\pm0.20$\% & $68.40\pm0.20$\% & $65.3\pm0.2$\% & $79.4\pm0.1$\%\\ 
Classical & R2-D2 & $48.39\pm0.29$\% & $68.24\pm0.26$\% & $62.9\pm0.3$\% & $82.8\pm0.3$\% \\ 
Classical w/ $R_{FC}$ & R2-D2 & $50.39\pm0.30$\% & ${\bf 69.58} \pm0.26$\% & ${\bf 65.5}\pm0.4$\% & ${\bf 83.3}\pm0.3$\%\\ 
Classical w/ $R_{HV}$ & R2-D2 & $50.16\pm0.30\%$ & $69.54\pm0.26\%$ & $64.6\pm0.3$\% & $83.1\pm0.3$\%\\ \hline 
MetaOptNet-SVM & MetaOptNet & ${\bf 62.64}\pm0.31$\% & ${\bf 78.63}\pm0.25$\% & $72.0\pm0.4$\% & $84.2\pm0.3$\%\\ 
Classical & MetaOptNet & $56.18\pm0.31$\% & $76.72\pm0.24$\% & $69.5\pm0.3$\% & $85.7\pm0.2$\% \\ 
Classical w/ $R_{FC}$ & MetaOptNet & $59.38\pm0.31$\% & $78.15\pm0.24$\% & ${\bf 72.3}\pm0.4$\% & ${\bf 86.3}\pm0.2$\%\\ 
Classical w/ $R_{HV}$ & MetaOptNet & $59.37\pm0.32$\% & $77.05\pm0.25$\% & $72.0\pm0.4$\% & $85.9\pm0.2$\%\\
\hline

\end{tabular}
\end{center}
\caption{Comparison of methods on 1-shot and 5-shot CIFAR-FS and mini-ImageNet 5-way classification. The top accuracy for each backbone/task is in bold.  Confidence intervals have radius equal to one standard error.  Few-shot fine-tuning is performed with SVM except for R2-D2, for which we report numbers from the original paper.}
\label{Feature_Space_Reg}
\end{table*}

\par In addition to performance evaluations, we calculate the similarity between feature representations yielded by a feature extractor produced by meta-learning and that of one produced by the classical routine with and without $R_{FC}$.  To this end, we use centered kernel alignment (CKA)~\cite{kornblith2019similarity}. Using both R2-D2 and MetaOptNet backbones on both mini-ImageNet and CIFAR-FS datasets, networks trained with $R_{FC}$ exhibit higher similarity scores to meta-learned networks than networks trained classically but without $R_{FC}$.  These measurements provide further evidence that feature clustering makes feature representations closer to those trained by meta-learning and thus, that meta-learners perform feature clustering. See Table \ref{tab:feature_similarity} for more details.

\begin{table}[h!]
\begin{center}
\begin{tabular}{|c|c|c|c|c|c|}
\hline
Backbone & Dataset & C & $R_{FC}$ & $R_{HV}$ \\ \hline
R2-D2 & CIFAR-FS& 0.71 & {\bf 0.77} & 0.73 \\ \hline
MetaOptNet & CIFAR-FS & 0.77 & {\bf 0.89} & 0.87 \\ \hline
R2-D2 & mini-ImageNet & 0.69 & {\bf 0.72} & 0.70 \\ \hline
MetaOptNet & mini-ImageNet & 0.70 & {\bf 0.82} & 0.79 \\ \hline

\end{tabular}
\end{center}
\caption{Similarity (CKA) representations trained via meta-learning and via transefer learning with/without the two proposed regularizers for various backbones and both CIFAR-FS and mini-ImageNet datasets. ``C'' denotes the classical transfer learning without regularizers. The highest score for each dataset/backbone combination is in bold.}
\label{tab:feature_similarity}
\end{table}

\subsection{Connecting Feature Clustering with Hyperplane Invariance}
\label{HyperplaneInvariance}

\par For further validation of the connection between feature clustering and invariance of separating hyperplanes to data sampling, we replace the feature clustering regularizer with one that penalizes variations in the maximum-margin hyperplane separating feature vectors in opposite classes.  Consider data points $x_1, x_2$ in class $A$, data points $y_1, y_2$ in class $B$, and feature extractor $f_\theta$.  The difference vector $f_\theta(x_1)-f_\theta(y_1)$ determines the direction of the maximum margin hyperplane separating the two points in feature space.  To penalize the variation in hyperplanes, we introduce the hyperplane variation regularizer,
\begin{multline*}
R_{HV}(f_\theta(x_1),f_\theta(x_2),f_\theta(y_1),f_\theta(y_2)) \\
=\frac{\|(f_\theta(x_1)-f_\theta(y_1))-(f_\theta(x_2)-f_\theta(y_2))\|_2}{\|(f_\theta(x_1)-f_\theta(y_1)\|_2+\|f_\theta(x_2)-f_\theta(y_2)\|_2}.
\end{multline*}

\par This function measures the distance between distance vectors $x_1-y_1$ and $x_2-y_2$ relative to their size.  In practice, during a batch of training, we sample many pairs of classes and two samples from each class.  Then, we compute $R_{HV}$ on all class pairs and add these terms to the cross-entropy loss.  We find that this regularizer performs almost as well as $R_{FC}$ and conclusively outperforms non-regularized classical training. We include these results in Table \ref{Feature_Space_Reg}.  See Appendix~\ref{sec:FeatureExpts} for more details on these experiments, including training times (which, as indicated in Section \ref{Regularizer1}, are significantly lower than those needed for meta-learning).

\subsection{MAML Does Not Have the Same Feature Separation Properties}
\label{MAML}

\par Remember that the previous measurements and experiments examined meta-learning methods which fix the feature extractor during the inner loop.  MAML is a popular example of a method which does not fix the feature extractor in the inner loop.  We now quantify MAML's class separation compared to transfer learning by computing our regularizer values for a pre-trained MAML model as well as a classically trained model of the same architecture.  We find that, in fact, MAML exhibits even worse feature separation than a classically trained model of the same architecture.  See Table \ref{MAMLClustering} for numerical results.  These results confirm our suspicion that the feature clustering phenomenon is specific to meta-learners which fix the feature extractor during the inner loop of training.

\begin{table}[h!]
\begin{center}
\begin{tabular}{|c|c|c|}
\hline
Model & $R_{FC}$ & $R_{HV}$ \\ \hline
MAML-1 & 3.9406 & 1.9434 \\ %\hline
MAML-5 & 3.7044 & 1.8901 \\ %\hline
MAML-C & \textbf{3.3487} & \textbf{1.8113} \\ \hline

\end{tabular}
\end{center}
\caption{Comparison of regularizer values for 1-shot and 5-shot MAML models (MAML-1 and MAML-5) as well as MAML-C, a classically trained model of the same architecture on mini-ImageNet training data.  The lowest value of each regularizer is in bold.}
\label{MAMLClustering}
\end{table}

\section{Finding Clusters of Local Minima for Task Losses in Parameter Space}
\label{ParameterClustering}

\par Since Reptile does not fix the feature extractor during fine-tuning, it must find parameters that adapt easily to new tasks.  One way Reptile might achieve this is by finding parameters that can reach a task-specific minimum by traversing a smooth, nearly linear region of the loss landscape.  In this case, even a single SGD update would move parameters in a useful direction. Unlike MAML, however, Reptile does not backpropagate through optimization steps and thus lacks information about the loss surface geometry when performing parameter updates.  Instead, we hypothesize that Reptile finds parameters that lie very close to good minima for many tasks and is therefore able to perform well on these tasks after very little fine-tuning.

This hypothesis is further motivated by the close relationship between Reptile and {\em consensus optimization} \cite{boyd2011distributed}.  In a consensus method, a number of models are independently optimized with their own task-specific parameters, and the tasks communicate via a penalty that encourages all the individual solutions to converge around a common value.  Reptile can be interpreted as approximately minimizing the consensus formulation
 $$   \frac{1}{m}\sum_{p=1}^m\mathcal{L}_{\mathcal{T}_p}(\tilde{\theta}_p) + \frac{\gamma}{2}  \| \tilde{\theta}_p - \theta\|^2,  $$
 where $\mathcal{L}_{\mathcal{T}_p}(\tilde{\theta}_p)$ is the loss for task $\mathcal{T}_p$, $\{\tilde{\theta}_p\}$ are task-specific parameters, and the quadratic penalty on the right encourages the parameters to cluster around a ``consensus value'' $\theta$.  A stochastic optimizer for this loss would proceed by alternately selecting a random task/term index $p$, minimizing the loss with respect to $\tilde{\theta}_p,$ and then taking a gradient step $\theta \gets \theta - \eta \tilde{\theta}_p$ to minimize the loss for $\theta.$  
 
 Reptile diverges from a traditional consensus optimizer only in that it does not explicitly consider the quadratic penalty term when minimizing for $\tilde{\theta}_p.$  However, it implicitly considers this penalty by initializing the optimizer for the task-specific loss using the current value of the consensus variables $\theta,$ which encourages the task-specific parameters to stay near the consensus parameters.  In the next section, we replace the standard Reptile algorithm with one that explicitly minimizes a consensus formulation.

\iffalse
\begin{itemize}
    \item Numerical measurements (Meta vs Transfer learning).
    \item Visualizations (scatter plots in parameter space?, histogram of distance traveled)
    \item Show that the same trends do not hold for MetaOptNet/R2-D2 with the histogram plot.
    \item Note on measuring distances using filter normalization.
\end{itemize}
\fi

\subsection{Consensus Optimization Improves Reptile}
\label{Reptile}

\par To validate the weight-space clustering hypothesis, we modify Reptile to explicitly enforce parameter clustering around a consensus value.  We find that directly optimizing the consensus formulation leads to improved performance.  To this end, during each inner loop update step in Reptile, we penalize the squared $\ell_2$ distance from the parameters for the current task to the average of the parameters across all tasks in the current batch.  Namely, we let:
$$R_i\big (\{\tilde{\theta}_p \}_{p=1}^m\big ) = d \big(\tilde{\theta}_i, \frac{1}{m}\sum_{p=1}^{m} \tilde{\theta}_p\big)^2 ,$$
where $\tilde{\theta}_p$ are the network parameters on task $p$ and $d$ is the filter normalized $\ell_2$ distance (see Note \ref{FilterNorm}).  Note that as parameters shrink towards the origin, the distances between minima shrink as well.  Thus, we employ filter normalization to ensure that our calculation is invariant to scaling \cite{li2018visualizing}.  See below for a description of filter normalization.  
This regularizer guides optimization to a location where many task-specific minima lie in close proximity.   A detailed description is given in Algorithm \ref{alg:RegReptile}, which is equivalent to the original Reptile when $\alpha=0$.  We call this method ``Weight-Clustering.''

\begin{note}
\label{FilterNorm}
Consider that a perturbation to the parameters of a network is more impactful when the network has small parameters. While previous work has used layer normalization or even more coarse normalization schemes, the authors of \citet{li2018visualizing} note that since the output of networks with batch normalization is invariant to filter scaling as long as the batch statistics are updated accordingly, we can normalize every filter of such a network independently.  The latter work suggests that this scheme, ``filter normalization'', correlates better with properties of the optimization landscape.  Thus, we measure distance in our regularizer using filter normalization, and we find that this technique prevents parameters from shrinking towards the origin.
\end{note}

\begin{algorithm}[h]
   \caption{Reptile with Weight-Clustering Regularization}
   \label{alg:RegReptile}
\begin{algorithmic}
\STATE {\bfseries Require:} Initial parameter vector, 
 $\theta$, outer learning rate, $\gamma$, inner learning rate, $\eta$, regularization coefficient, $\alpha$, and distribution over tasks, $p(\mathcal{T})$. 
\FOR{\textit{meta-step} $=1,\dots,n$}
\STATE Sample batch of tasks, $\{\mathcal{T}_i\}_{i=1}^m$ from $p(\mathcal{T})$\\
\vspace{1mm}
\STATE Initialize parameter vectors $\tilde{\theta}_i^0 = \theta$ for each task
\FOR{$j=1,\dots,k$}
\FOR{$i=1,\dots,m$}
\STATE Calculate $\mathcal{L} = \mathcal{L}_{\mathcal{T}_i}^j + \alpha R_i\big (\{\tilde{\theta}_p^{j-1} \}_{p=1}^m\big )$ 
\STATE Update $\tilde{\theta}_i^j = \tilde{\theta}_i^{j-1} - \eta \nabla_{\tilde{\theta}_i}  \mathcal{L}$
\ENDFOR
\ENDFOR
\STATE Compute difference vectors $\{g_i = \tilde{\theta}_i^k - \tilde{\theta}_i^0\}_{i=1}^m$
\STATE Update $\theta \leftarrow \theta - \frac{\gamma}{m} \sum_i g_i$
\ENDFOR
\end{algorithmic}
\end{algorithm}

\par We compare the performance of our regularized Reptile algorithm to that of the original Reptile method as well as first-order MAML (FOMAML) and a classically trained model of the same architecture.  We test these methods on a sample of 100,000 5-way 1-shot and 5-shot mini-ImageNet tasks and find that in both cases, Reptile with Weight-Clustering achieves higher performance than the original algorithm and significantly better performance than FOMAML and the classically trained models. These results are summarized in Table \ref{Clustering_MI}. 

\begin{table}[h]
\begin{center}
\begin{tabular}{|l|c|c|}
\hline
Framework & 1-shot & 5-shot \\ \hline
Classical & $28.72\pm0.16$\% & $45.25\pm0.21$\% \\ \hline
FOMAML & $48.07\pm1.75$\% & $63.15\pm0.91$\% \\ \hline
Reptile & $49.97\pm0.32$\% & $65.99\pm0.58$\% \\ \hline
W-Clustering & ${\bf 51.94} \pm0.23$\% & ${\bf 68.02}\pm0.22$\% \\ \hline

\end{tabular}
\end{center}
\caption{Comparison of methods on 1-shot and 5-shot mini-ImageNet 5-way classification.  The top accuracy for each task is in bold.  Confidence intervals have width equal to one standard error.  W-Clustering denotes the Weight-Clustering regularizer.}
\label{Clustering_MI}
\end{table}

\par We note that the best-performing result was attained when the product of the constant term collected from the gradient of the regularizer $R_i$ and the regularization coefficient $\alpha$ was $5.0\times10^{-5}$, but a range of values up to ten times larger and smaller also produced improvements over the original algorithm. Experimental details, as well as results for other values of this coefficient, can be found in Appendix \ref{sec:WeightExpts}.

\par
\begin{figure}[h]
    \centering
    \subfloat[]{{\includegraphics[width=\columnwidth]{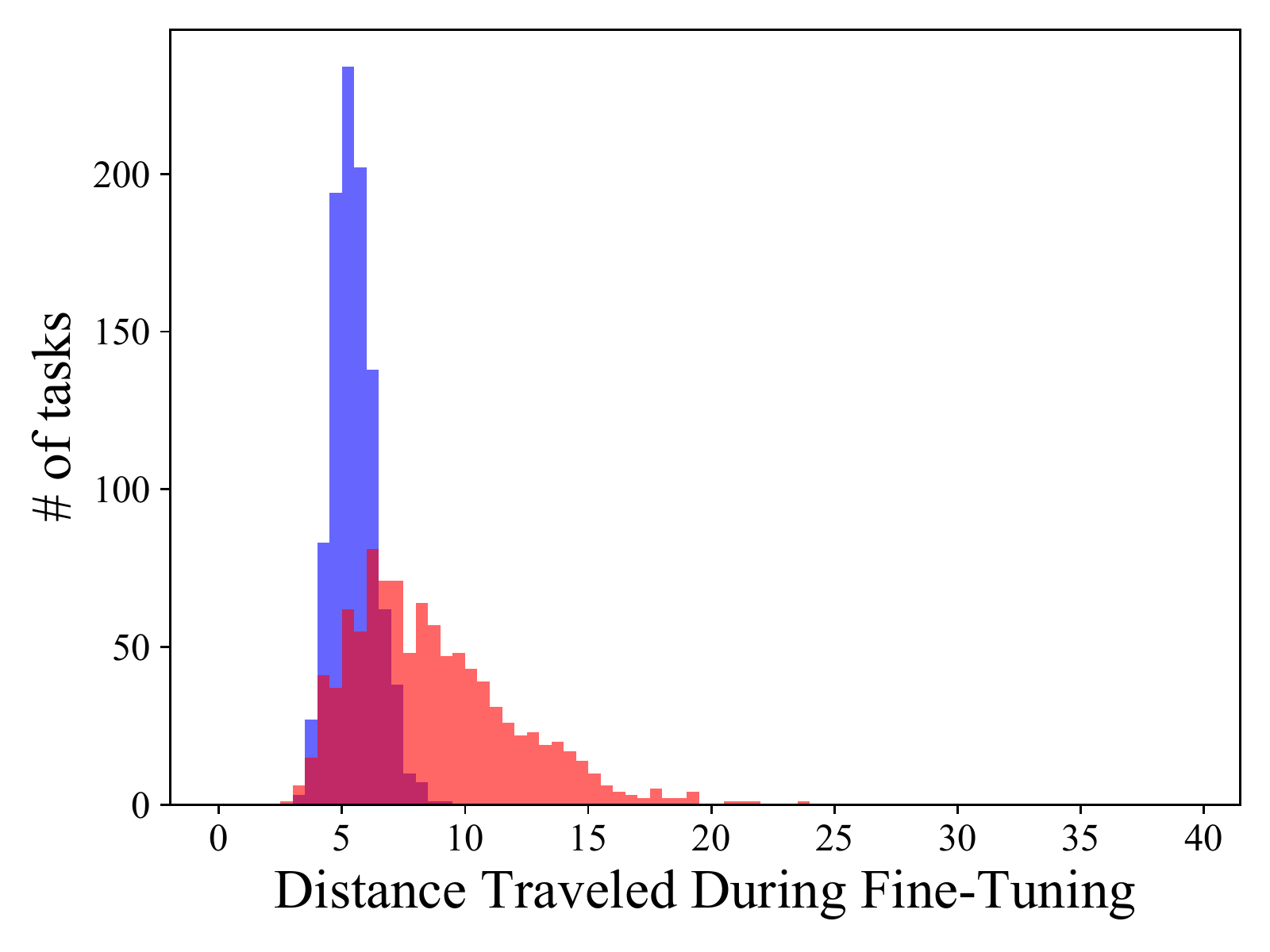} }}
    \qquad
    \subfloat[]{{\includegraphics[width=\columnwidth]{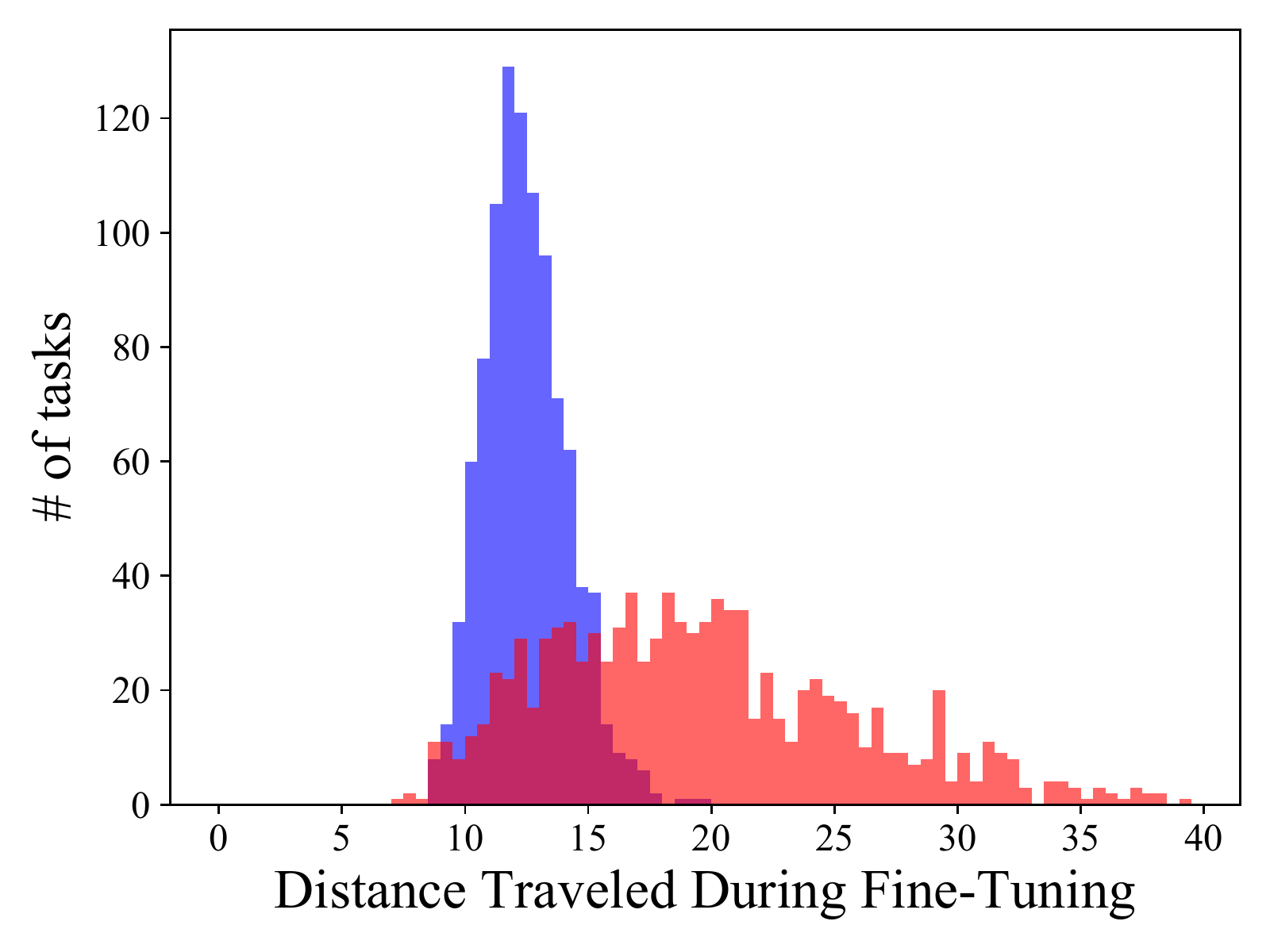} }}
    \caption{Histogram of filter normalized distance traveled during fine-tuning on a) 1-shot and b) 5-shot mini-ImageNet tasks by models trained using vanilla Reptile (red) and weight-clustered Reptile (blue).}
    \label{fig:DistHist}
\end{figure}

 In addition to these performance gains, we found that the parameters of networks trained using our regularized version of Reptile do not travel as far during fine-tuning at inference as those trained using vanilla Reptile. Figure \ref{fig:DistHist} depicts histograms of filter normalized distance traveled by both networks fine-tuning on samples of 1,000 1-shot and 5-shot mini-ImageNet tasks. From these, we conclude that our regularizer does indeed move model parameters toward a consensus which is near good minima for many tasks.  Interestingly, we applied these same measurements to networks trained using MetaOptNet and R2-D2, and we found that these feature extractors lie in wide and flat minimizers across many task losses.  Thus, when the whole network is fine-tuned, the parameters move a lot without substantially decreasing loss.  Previous work has associated flat minimizers with good generalization \cite{huang2019understanding}.

\iffalse

\begin{table}[h!]
\begin{center}
\begin{tabular}{|l|c|c|}
\hline
Framework & 1-shot & 5-shot \\ \hline
Transfer & $66.21\pm0.20\%$ & $89.56\pm0.14\%$ \\ \hline
FOMAML & ${\bf 98.30}\pm0.50$\% & $99.20\pm0.20$\% \\ \hline
Reptile & $97.68\pm0.04$\% & $99.48\pm0.06$\% \\ \hline
W-Clustering & $97.63\pm0.07\%$ & ${\bf 99.60}\pm0.03\%$ \\ \hline

\end{tabular}
\end{center}
\caption{Comparison of different methods on 1-shot and 5-shot Omniglot 5-way classification.  The top accuracy for each setup is in bold.}
\label{Clustering_Omni}
\end{table}

\fi

\section{Discussion}
\label{Discussion}

\par In this work, we shed light on two key differences between meta-learned networks and their classically trained counterparts.  We find evidence that meta-learning algorithms minimize the variation between feature vectors within a class relative to the variation between classes.  Moreover, we design two regularizers for transfer learning inspired by this principal, and our regularizers consistently improve few-shot performance.  The success of our method helps to confirm the hypothesis that minimizing within-class feature variation is critical for few-shot performance.

\par We further notice that Reptile resembles a consensus optimization algorithm, and we enhance the method by designing yet another regularizer, which we apply to Reptile, in order to find clusters of local minima in the loss landscapes of tasks.  We find in our experiments that this regularizer improves both one-shot and five-shot performance of Reptile on mini-ImageNet.

\par A PyTorch implementation of the feature clustering and hyperplane variation regularizers can be found at:\\
\url{https://github.com/goldblum/FeatureClustering}

\section*{Acknowledgements}
This work was supported by the ONR MURI program, the DARPA YFA program, DARPA GARD, the JHU HLTCOE, and the National Science Foundation DMS division.

\bibliography{refs}

\begin{thebibliography}{25}
\providecommand{\natexlab}[1]{#1}
\providecommand{\url}[1]{\texttt{#1}}
\expandafter\ifx\csname urlstyle\endcsname\relax
  \providecommand{\doi}[1]{doi: #1}\else
  \providecommand{\doi}{doi: \begingroup \urlstyle{rm}\Url}\fi

\bibitem[Altae-Tran et~al.(2017)Altae-Tran, Ramsundar, Pappu, and
  Pande]{altae2017low}
Altae-Tran, H., Ramsundar, B., Pappu, A.~S., and Pande, V.
\newblock Low data drug discovery with one-shot learning.
\newblock \emph{ACS central science}, 3\penalty0 (4):\penalty0 283--293, 2017.

\bibitem[Bertinetto et~al.(2018)Bertinetto, Henriques, Torr, and
  Vedaldi]{bertinetto2018meta}
Bertinetto, L., Henriques, J.~F., Torr, P.~H., and Vedaldi, A.
\newblock Meta-learning with differentiable closed-form solvers.
\newblock \emph{arXiv preprint arXiv:1805.08136}, 2018.

\bibitem[Boyd et~al.(2011)Boyd, Parikh, Chu, Peleato, Eckstein,
  et~al.]{boyd2011distributed}
Boyd, S., Parikh, N., Chu, E., Peleato, B., Eckstein, J., et~al.
\newblock Distributed optimization and statistical learning via the alternating
  direction method of multipliers.
\newblock \emph{Foundations and Trends{\textregistered} in Machine learning},
  3\penalty0 (1):\penalty0 1--122, 2011.

\bibitem[Chen et~al.(2019)Chen, Liu, Kira, Wang, and Huang]{chen2019closer}
Chen, W.-Y., Liu, Y.-C., Kira, Z., Wang, Y.-C.~F., and Huang, J.-B.
\newblock A closer look at few-shot classification.
\newblock \emph{arXiv preprint arXiv:1904.04232}, 2019.

\bibitem[Dhillon et~al.(2019)Dhillon, Chaudhari, Ravichandran, and
  Soatto]{dhillon2019baseline}
Dhillon, G.~S., Chaudhari, P., Ravichandran, A., and Soatto, S.
\newblock A baseline for few-shot image classification.
\newblock \emph{arXiv preprint arXiv:1909.02729}, 2019.

\bibitem[Finn et~al.(2017)Finn, Abbeel, and Levine]{finn2017model}
Finn, C., Abbeel, P., and Levine, S.
\newblock Model-agnostic meta-learning for fast adaptation of deep networks.
\newblock In \emph{Proceedings of the 34th International Conference on Machine
  Learning-Volume 70}, pp.\  1126--1135. JMLR. org, 2017.

\bibitem[Frosst et~al.(2019)Frosst, Papernot, and Hinton]{frosst2019analyzing}
Frosst, N., Papernot, N., and Hinton, G.
\newblock Analyzing and improving representations with the soft nearest
  neighbor loss.
\newblock \emph{arXiv preprint arXiv:1902.01889}, 2019.

\bibitem[Goldblum et~al.(2019{\natexlab{a}})Goldblum, Fowl, and
  Goldstein]{goldblum2019robust}
Goldblum, M., Fowl, L., and Goldstein, T.
\newblock Robust few-shot learning with adversarially queried meta-learners.
\newblock \emph{arXiv preprint arXiv:1910.00982}, 2019{\natexlab{a}}.

\bibitem[Goldblum et~al.(2019{\natexlab{b}})Goldblum, Geiping, Schwarzschild,
  Moeller, and Goldstein]{goldblum2019truth}
Goldblum, M., Geiping, J., Schwarzschild, A., Moeller, M., and Goldstein, T.
\newblock Truth or backpropaganda? an empirical investigation of deep learning
  theory.
\newblock In \emph{International Conference on Learning Representations},
  2019{\natexlab{b}}.

\bibitem[He et~al.(2016)He, Zhang, Ren, and Sun]{he2016deep}
He, K., Zhang, X., Ren, S., and Sun, J.
\newblock Deep residual learning for image recognition.
\newblock In \emph{Proceedings of the IEEE conference on computer vision and
  pattern recognition}, pp.\  770--778, 2016.

\bibitem[Huang et~al.(2019{\natexlab{a}})Huang, Larochelle, and
  Lacoste{-}Julien]{huang2019centroid}
Huang, G., Larochelle, H., and Lacoste{-}Julien, S.
\newblock Centroid networks for few-shot clustering and unsupervised few-shot
  classification.
\newblock \emph{CoRR}, abs/1902.08605, 2019{\natexlab{a}}.
\newblock URL \url{http://arxiv.org/abs/1902.08605}.

\bibitem[Huang et~al.(2019{\natexlab{b}})Huang, Emam, Goldblum, Fowl, Terry,
  Huang, and Goldstein]{huang2019understanding}
Huang, W.~R., Emam, Z., Goldblum, M., Fowl, L., Terry, J.~K., Huang, F., and
  Goldstein, T.
\newblock Understanding generalization through visualizations.
\newblock \emph{arXiv preprint arXiv:1906.03291}, 2019{\natexlab{b}}.

\bibitem[Kornblith et~al.(2019{\natexlab{a}})Kornblith, Norouzi, Lee, and
  Hinton]{kornblith2019similarity}
Kornblith, S., Norouzi, M., Lee, H., and Hinton, G.
\newblock Similarity of neural network representations revisited.
\newblock \emph{arXiv preprint arXiv:1905.00414}, 2019{\natexlab{a}}.

\bibitem[Kornblith et~al.(2019{\natexlab{b}})Kornblith, Shlens, and
  Le]{kornblith2019better}
Kornblith, S., Shlens, J., and Le, Q.~V.
\newblock Do better imagenet models transfer better?
\newblock In \emph{Proceedings of the IEEE conference on computer vision and
  pattern recognition}, pp.\  2661--2671, 2019{\natexlab{b}}.

\bibitem[Lee et~al.(2019)Lee, Maji, Ravichandran, and Soatto]{lee2019meta}
Lee, K., Maji, S., Ravichandran, A., and Soatto, S.
\newblock Meta-learning with differentiable convex optimization.
\newblock In \emph{Proceedings of the IEEE Conference on Computer Vision and
  Pattern Recognition}, pp.\  10657--10665, 2019.

\bibitem[Li et~al.(2018)Li, Xu, Taylor, Studer, and
  Goldstein]{li2018visualizing}
Li, H., Xu, Z., Taylor, G., Studer, C., and Goldstein, T.
\newblock Visualizing the loss landscape of neural nets.
\newblock In \emph{Advances in Neural Information Processing Systems}, pp.\
  6389--6399, 2018.

\bibitem[Mika et~al.(1999)Mika, Ratsch, Weston, Scholkopf, and
  Mullers]{mika1999fisher}
Mika, S., Ratsch, G., Weston, J., Scholkopf, B., and Mullers, K.-R.
\newblock Fisher discriminant analysis with kernels.
\newblock In \emph{Neural networks for signal processing IX: Proceedings of the
  1999 IEEE signal processing society workshop (cat. no. 98th8468)}, pp.\
  41--48. Ieee, 1999.

\bibitem[Nagabandi et~al.(2018)Nagabandi, Clavera, Liu, Fearing, Abbeel,
  Levine, and Finn]{nagabandi2018learning}
Nagabandi, A., Clavera, I., Liu, S., Fearing, R.~S., Abbeel, P., Levine, S.,
  and Finn, C.
\newblock Learning to adapt in dynamic, real-world environments through
  meta-reinforcement learning.
\newblock \emph{arXiv preprint arXiv:1803.11347}, 2018.

\bibitem[Nichol \& Schulman(2018)Nichol and Schulman]{nichol2018reptile}
Nichol, A. and Schulman, J.
\newblock Reptile: a scalable metalearning algorithm.
\newblock \emph{arXiv preprint arXiv:1803.02999}, 2:\penalty0 2, 2018.

\bibitem[Oreshkin et~al.(2018)Oreshkin, L{\'o}pez, and
  Lacoste]{oreshkin2018tadam}
Oreshkin, B., L{\'o}pez, P.~R., and Lacoste, A.
\newblock Tadam: Task dependent adaptive metric for improved few-shot learning.
\newblock In \emph{Advances in Neural Information Processing Systems}, pp.\
  721--731, 2018.

\bibitem[Sainath et~al.(2013)Sainath, Kingsbury, Sindhwani, Arisoy, and
  Ramabhadran]{sainath2013low}
Sainath, T.~N., Kingsbury, B., Sindhwani, V., Arisoy, E., and Ramabhadran, B.
\newblock Low-rank matrix factorization for deep neural network training with
  high-dimensional output targets.
\newblock In \emph{2013 IEEE international conference on acoustics, speech and
  signal processing}, pp.\  6655--6659. IEEE, 2013.

\bibitem[Snell et~al.(2017)Snell, Swersky, and Zemel]{snell2017prototypical}
Snell, J., Swersky, K., and Zemel, R.
\newblock Prototypical networks for few-shot learning.
\newblock In \emph{Advances in Neural Information Processing Systems}, pp.\
  4077--4087, 2017.

\bibitem[Song et~al.(2019)Song, Liu, and Qin]{song2019fast}
Song, L., Liu, J., and Qin, Y.
\newblock Fast and generalized adaptation for few-shot learning.
\newblock \emph{arXiv preprint arXiv:1911.10807}, 2019.

\bibitem[Vinyals et~al.(2016)Vinyals, Blundell, Lillicrap, Wierstra,
  et~al.]{vinyals2016matching}
Vinyals, O., Blundell, C., Lillicrap, T., Wierstra, D., et~al.
\newblock Matching networks for one shot learning.
\newblock In \emph{Advances in neural information processing systems}, pp.\
  3630--3638, 2016.

\bibitem[Wang et~al.(2020)Wang, Gao, Zhao, Li, Dou, and Xu]{wang2020pay}
Wang, K., Gao, X., Zhao, Y., Li, X., Dou, D., and Xu, C.-Z.
\newblock Pay attention to features, transfer learn faster {CNN}s.
\newblock In \emph{International Conference on Learning Representations}, 2020.
\newblock URL \url{https://openreview.net/forum?id=ryxyCeHtPB}.

\end{thebibliography}
\bibliographystyle{icml2020}

\clearpage
\appendix
\onecolumn
\section{Experimental Details}\label{sec:exp}

The mini-ImageNet and CIFAR-FS datasets can be found at \url{https://github.com/yaoyao-liu/mini-imagenet-tools} and \url{https://github.com/ArnoutDevos/maml-cifar-fs respectively}.

\subsection{Mixing Meta-Learned Models and Fine-Tuning Procedures: Additional Experiments}
\label{sec:ModelSwap}

\begin{table}[h]
\begin{center}
\begin{tabular}{|c|c|c|c|c|}
\hline
Model & SVM & RR & ProtoNet & MAML \\ \hline
MetaOptNet-M & \textbf{78.63} $\pm$ 0.25 \% & \textbf{76.96} $\pm$ 0.23 \% & \textbf{76.17} $\pm$ 0.23 \% & 70.14 $\pm$ 0.27 \% \\ %\hline
MetaOptNet-C & 76.72 $\pm$ 0.24 \% & 74.48 $\pm$ 0.24 \% & 73.37 $\pm$ 0.24 \% & \textbf{71.32} $\pm$ 0.26 \% \\ \hline
R2-D2-M & \textbf{68.40} $\pm$ 0.20 \% & \textbf{72.09} $\pm$ 0.25 \% & \textbf{70.74} $\pm$ 0.25 \% & \textbf{71.43} $\pm$ 0.27 \% \\ %\hline
R2-D2-C & 68.24 $\pm$ 0.26 \% & 67.04 $\pm$ 0.26 \% & 60.93 $\pm$ 0.29 \% & 65.30 $\pm$ 0.27 \% \\ \hline
\end{tabular}
\end{center}
\caption{Comparison of meta-learning and transfer learning models with various fine-tuning algorithms on 5-shot mini-ImageNet.  ``MetaOptNet-M'' and ``MetaOptNet-C'' denote models with MetaOptNet backbone trained with MetaOptNet-SVM and classical training.  Similarly, ``R2-D2-M'' and ``R2-D2-C'' denote models with R2-D2 backbone trained with ridge regression (RR) and classical training.  Column headers denote the fine-tuning algorithm used for evaluation, and the radius of confidence intervals is one standard error.}
\label{MvT5MI}
\end{table}

\subsection{Transfer Learning and Feature Space Clustering}
\label{sec:FeatureExpts}
\par We evaluate the proposed regularizers and classically trained baseline on two backbone architectures: a 4-layer convolutional neural network with number of filters per layer 96-192-384-512 originally used for R2-D2 \cite{bertinetto2018meta} and ResNet-12~\cite{he2016deep,oreshkin2018tadam,lee2019meta}. We run experiments on the mini-ImageNet and CIFAR-FS datasets.

\par When training the backbone feature extractors, we use SGD with a batch-size of 128 for CIFAR-FS and 256 for mini-ImageNet, Nesterov momentum set to 0.9 and weight decay of $10^{-4}$. For training on CIFAR-FS, we set the initial learning rate to 0.1 for the first 100 epochs and reduce by a factor of 10 every 50 epochs. To avoid gradient explosion problems, we use 15 warm-up epochs for mini-ImageNet with learning rate 0.01.  We train all classically trained networks for a total of 300 epochs. We employ data parallelism across 2 Nvidia RTX 2080 Ti GPUs when training on mini-ImageNet, and we only use one GPU for each CIFAR-FS experiment. For few-shot testing, we train two classification heads, a linear NN layer and SVM ~\cite{lee2019meta} on top of the pre-trained feature extractors. The evaluation results of these models are given in Table~\ref{Feature_Space_Reg_Hyper_Param}. Table~\ref{Feature_Space_Reg_Runtime} shows the running time per training epoch as well as total training time on both datasets and backbone architectures to achieve the results in Table~\ref{Feature_Space_Reg}. The training speed of the proposed regularizers is nearly as fast as classical transfer learning and up to almost 13 times faster than meta-learning methods. For meta-learning methods, we follow the training hyperparemeters from \cite{lee2019meta}.

\begin{table}[h!]
\begin{center}
\begin{tabular}{lccc}
\hline
 & & mini-ImageNet& CIFAR-FS\\\cline{3-4}
Training & Backbone & runtime & runtime\\ \hline
R2-D2 & R2-D2 & 16m/16.8h & 44s/45m\\ 
Classical & R2-D2 & 20s/1.7h & 4s/22m \\ 
Classical w/ $R_{FC}$ & R2-D2 & 20s/1.7h & 4s/24m \\ 
Classical w/ $R_{HV}$ & R2-D2 & 20s/1.7h & 4s/23m\\ \hline 
MetaOptNet-SVM & MetaOptNet & 1.5h/88.0h & 4m/4.5h \\ 
Classical & MetaOptNet &1.4m/7.0h & 14s/1.2h \\ 
Classical w/ $R_{FC}$ & MetaOptNet & 1.5m/7.4h & 15s/1.3h\\ 
Classical w/ $R_{HV}$ & MetaOptNet & 1.3m/7.2h & 16s/1.4h\\
\hline

\end{tabular}
\end{center}
\caption{Runtime (training time per epoch/total times) comparison of methods on CIFAR-FS and mini-ImageNet 5-way classification on a single GPU.}
\label{Feature_Space_Reg_Runtime}
\end{table}

\begin{table*}[h!]
\begin{center}
\begin{tabular}{lccccccc}
\hline
& & & & \multicolumn{2}{c}{mini-ImageNet}&\multicolumn{2}{c}{CIFAR-FS}\\\cline{5-8}
Backbone & Regularizer & Coeff & Head  & 1-shot & 5-shot & 1-shot & 5-shot\\ \hline
R2-D2 & $R_{FC}$ & 0.02  & NN &$48.27\pm 0.29$\% & $69.13\pm0.26$\% & $63.11\pm0.35$\% & $83.31\pm0.25$\%\\ 
& & 0.05 & NN& $48.75\pm0.29$\% & $69.50\pm0.26$\% & $64.49\pm0.35$\% & $83.32\pm0.25$\% \\ 
& & 0.1 & NN& $48.72\pm0.29$\% & $67.39\pm0.25$\% & $62.98\pm0.36$\% & $81.07\pm0.26$\%\\
& $R_{HV}$& 0.02  & NN & $46.74\pm0.28$\% & $68.19\pm0.27$\% & $62.50\pm0.34$\% & $82.90\pm0.25$\%\\ 
& & 0.05 & NN& $49.11\pm0.29$\% & $68.88\pm0.26$\% & $63.61\pm0.35$\% & $83.21\pm0.25$ \%\\ 
& & 0.1 & NN& $48.87\pm0.29$\% & $69.67\pm0.26$\% & $63.50\pm0.35$\% & $83.17\pm0.25$\%\\
& $R_{FC}$ & 0.02 & SVM &$49.05\pm0.30$\% & $68.94\pm0.26$\% & $64.48\pm0.34$\% & $83.11\pm0.25$\%\\ 
& & 0.05 & SVM& $50.39\pm0.30$\% & $69.58\pm0.26$\% & $65.53\pm0.35$\% & $83.30\pm0.25$\% \\ 
& & 0.1 & SVM& $50.71\pm0.30$\% & $68.46\pm0.25$\% & $64.25\pm0.36$\% & $81.57\pm0.26$\%\\
& $R_{HV}$& 0.02  & SVM & $47.81\pm0.29$\% & $68.08\pm0.27$\% & $63.71\pm0.33$\% & $82.77\pm0.26$\%\\ 
& & 0.05 & SVM& $49.28\pm0.30$\% & $68.62\pm0.26$\% & $64.52\pm0.34$\% & $82.99\pm0.26$\%\\ 
& & 0.1 & SVM& $50.16\pm0.30\%$ & $69.54\pm0.26$\% & $64.62\pm0.34$\% & $83.08\pm0.26$\%\\
ResNet-12 &$R_{FC}$& 0.02 & NN & $57.54\pm0.32$\% & $77.31\pm0.25$\% & $71.69\pm0.36$\% & $86.13\pm0.23$\%\\ 
& & 0.05 & NN & $56.59\pm0.33$\% & $74.81\pm0.25$\% & $71.78\pm0.37$\% & $85.30\pm 0.24$\%\\ 
& & 0.1 & NN & $52.26\pm0.35$\% & $69.93\pm0.28$\% & $71.85\pm0.39$\% & $83.74\pm0.25$\%\\
& $R_{HV}$& 0.02  & NN & $53.75\pm0.30$\% & $76.11\pm0.25$\% & $70.12\pm0.35$\% & $86.37\pm0.23$\%\\ 
& & 0.05 & NN& $57.15\pm0.31$\% & $77.27\pm0.25$\% & $71.49\pm0.36$\% & $85.85\pm0.24$\%\\ 
& & 0.1 & NN& $57.76\pm0.33\%$ & $76.05\pm0.26$\% & $71.56\pm0.37$\% & $84.80\pm0.25$\%\\
& $R_{FC}$ & 0.02 & SVM &$59.38\pm0.31$\% & $78.15\pm0.24$\% & $72.32\pm0.30$\% & $86.31\pm0.24$\%\\ 
& & 0.05 & SVM& $59.05\pm0.32$\% & $76.36\pm0.24$\% & $71.94\pm0.36$\% & $85.28\pm0.24$\% \\ 
& & 0.1 & SVM& $56.73\pm0.35$\% & $73.70\pm0.26$\% & $71.08\pm0.36$\% & $83.49\pm0.25$\%\\
& $R_{HV}$& 0.02  & SVM & $56.95\pm0.30$\% & $77.06\pm0.24$\% & $71.34\pm0.35$\% & $86.54\pm0.23$\%\\ 
& & 0.05 & SVM& $59.36\pm0.31$\% & $77.97\pm0.24$\% & $72.00\pm0.36$\% & $85.87\pm0.24$\%\\ 
& & 0.1 & SVM& $59.37\pm0.32$\% & $77.05\pm0.25$\% & $71.92\pm0.37$\% & $84.84\pm0.25$\%\\
\hline

\end{tabular}
\end{center}
\caption{Hyper-parameter tuning for $R_{FC}$ and $R_{HV}$ regularizers with various backbone structures and classification heads on 1-shot and 5-shot CIFAR-FS and mini-ImageNet 5-way classification.  Regularizer coefficients include the $C/N$ factor.}
\label{Feature_Space_Reg_Hyper_Param}
\end{table*}

\pagebreak

\subsection{Reptile Weight Clustering}
\label{sec:WeightExpts}

We train models via our weight-clustering Reptile algorithm with a range of coefficients for the regularization term. The model architecture and all other hyperparameters were chosen to match those specified for Reptile training and evaluation on 1-shot and 5-shot mini-ImageNet in \cite{nichol2018reptile}. The evaluation results of these models are given in Table \ref{ClusterCoeffs}.  All models were trained on Nvidia RTX 2080 Ti GPUs.

\begin{table}[h!]
\begin{center}
\renewcommand{\arraystretch}{1.1}
\begin{tabular}{|l|c|c|}
\hline
Coefficient & 1-shot & 5-shot \\ \hline
$0$ (Reptile) & $49.97\pm0.32$\% & $65.99\pm0.58$\% \\ \hline
$1.0\times10^{-5}$ & $51.42\pm0.23\%$ & $67.16\pm0.22\%$ \\ \hline
$2.5\times10^{-5}$ & $51.25\pm0.24\%$ & $67.55\pm0.22\%$ \\ \hline
$5.0\times10^{-5}$ & ${\bf 51.94} \pm0.23\%$ & ${\bf 68.02}\pm0.22\%$ \\ \hline
$7.5\times10^{-5}$ & $51.40\pm0.24\%$ & $67.59\pm0.22\%$ \\ \hline
$1.0\times10^{-4}$ & $50.92\pm0.23\%$ & $67.91\pm0.22\%$ \\ \hline
$2.5\times10^{-4}$ & $50.65\pm0.23\%$ & $65.95\pm0.23\%$ \\ \hline
$5.0\times10^{-4}$ & $51.37\pm0.23\%$ & $66.98\pm0.23\%$ \\ \hline

\end{tabular}
\end{center}
\caption{Comparison of test accuracy for models trained with the weight-clustering Reptile algorithm with various regularization coefficients evaluated on 1-shot and 5-shot mini-ImageNet tasks. The results for vanilla Reptile are those given in \cite{nichol2018reptile}.}
\label{ClusterCoeffs}
\end{table}

\subsection{Architectures}

For our experiments using MAML, R2-D2, MetaOptNet, and Reptile, we use the architectures originally used for experiments in the respective papers \cite{finn2017model, bertinetto2018meta, lee2019meta, nichol2018reptile}. Specificaly, \cite{finn2017model, nichol2018reptile} use the same network with $4$ convolutional layers. \cite{bertinetto2018meta} uses a modified version of this convolutional network, while \cite{lee2019meta} employs a ResNet-12 architecture. 
\section{Proof of Theorem 1}
\label{sec:Proof}

Consider the three conditions 
$$\|x-\overline X\| < \delta, \quad \|y-\overline  Y\| < \delta, \quad \|z- \overline X\| < \delta,$$
 where  $\delta = \|\overline X - \overline Y\|/4,$ and $\overline X$ is the expected value of $X.$ Under these conditions, 
 $$\|z-x\| \le \|z-\overline X\|+\|x-\overline X\| < 2 \delta $$
 and
 $$\|z-y\| \ge \|\overline X - \overline Y\| - \|y - \overline Y\| -  \|z - \overline X\| > 4\delta - 2\delta = 2\delta.$$
 Combining the above yields
$$\|z-x\|<\|z-y\|.$$ 
We can now write
\begin{align*}
z^T(x - y)& - \frac{1}{2}\|x\|^2 +\frac{1}{2}\|y\|^2 \\
%&= z^T(x - y)+\frac{1}{2}(y+x)^T(y-x) \\
%&= (z-x)^T(x - y)+x^T(x - y) \\
 % &\qquad+\frac{1}{2}(y+x)^T(y-x) \\
%&= (z-x)^T(x - y)+\frac{1}{2}(y-x)^T(y-x) \\
%&= (z-x)^T(x - y)+\frac{1}{2}\|y-x\|^2 \\
&=  -\|z-x\|^2+\frac{1}{2}\|z-y\|^2 +\frac{1}{2}\|z-x\|^2 \\
&\ge  -\|z-x\|^2+\frac{1}{2}\|z-x\|^2 +\frac{1}{2}\|z-x\|^2 \\
&= 0,
%&= [z-\frac{1}{2}(y+x) ] ^T(x-y)
\end{align*}
and so $z$ is classified correctly if our three conditions hold.   From the Chebyshev bound, these conditions hold with probability at least
 \begin{align}\label{cheb_bound}
  \left(1 - \frac{\sigma_x^2}{\delta^2}\right)^2\left(1 - \frac{\sigma_y^2}{\delta^2}\right) \ge 
   \left(1 - \frac{2\sigma_x^2}{\delta^2}\right)\left(1 - \frac{\sigma_y^2}{\delta^2}\right) \ge 
   1 - \frac{2\sigma_x^2+\sigma_y^2}{\delta^2},
   \end{align}
   where we have twice applied the identity $(1-a)(1-b)\ge (1-a-b),$ which holds for $a,b\ge0,$ (this also requires $\sigma_y^2/\delta^2<1$, but this can be guaranteed by choosing a sufficiently small $\epsilon$ as in the statement of the theorem). 

\par Finally, we have the variation ratio bound
$$ \frac{\text{var}[X]+\text{var}[Y]}{\text{var}[U]} =  \frac{\sigma_x^2+\sigma_y^2}{\sigma_x^2+\sigma_y^2+16\delta^2}  <\epsilon. $$
And so 
$$\delta^2 \ge  \frac{(1-\epsilon)(\sigma_x^2+\sigma_y^2)}{16\epsilon} .$$
Plugging this into \eqref{cheb_bound} we get the final probability bound
 \begin{align}
  1 - \frac{32\epsilon\sigma_x^2+16\epsilon\sigma_y^2}{(\sigma_x^2+\sigma_y^2)(1-\epsilon)} \ge 1 - \frac{32\epsilon\sigma_x^2+32\epsilon\sigma_y^2}{(\sigma_x^2+\sigma_y^2)(1-\epsilon)} =  1 - \frac{32\epsilon}{(1-\epsilon)}.
   \end{align}

\end{document}